\documentclass[lettersize,journal]{IEEEtran}

\usepackage{amsmath,amsfonts,amssymb,mathtools}
\usepackage{algorithmic}
\usepackage{algorithm}
\usepackage{array}
\usepackage{subcaption}
\usepackage{textcomp}
\usepackage{multirow}
\usepackage{stfloats}
\usepackage{url}
\usepackage{verbatim}
\usepackage{graphicx}
\usepackage{cite}
\usepackage{wrapfig}
\usepackage{booktabs}
\usepackage{nicefrac}
\usepackage{microtype}
\usepackage{tikz}
\usepackage{soul}
\usepackage{multicol}
\usepackage[linesnumbered,ruled,vlined,algo2e,algoruled,boxed]{algorithm2e}
\usepackage{etoolbox}
\usepackage[hidelinks]{hyperref}
\hypersetup{
    colorlinks=true,
    linkcolor=blue,
    filecolor=magenta,
    urlcolor=cyan,
    citecolor=red
}

\SetKwInput{KwInit}{Initialize}

\makeatletter
\patchcmd{\@begintheorem}{\textit}{\textbf}{}{}
\makeatother

\newcommand{\bG}{{\mathbf G}}

\newcommand{\bX}{{\mathbf X}}

\newcommand{\bL}{{\mathbf L}}

\begin{document}

\title{Efficient Reinforcement Learning using Linear Koopman Dynamics for Nonlinear Robotic Systems}

\author{Wenjian Hao, Yuxuan Fang, Zehui Lu, and Shaoshuai Mou
\thanks{W. Hao, Y. Fang, and S. Mou are with the School of Aeronautics and Astronautics, Purdue University, West Lafayette, IN 47907, USA. { (Email: \texttt{\{hao93, fang394, mous\}@purdue.edu}) }}
\thanks{Z. Lu is an independent researcher. (Email:  \texttt{zehuilu789@gmail.com})}
}

\maketitle

\begin{abstract}
This paper presents a model-based reinforcement learning (RL) framework for optimal closed-loop control of nonlinear robotic systems. The proposed approach learns linear lifted dynamics through Koopman operator theory and integrates the resulting model into an actor–critic architecture for policy optimization, where the policy represents a parameterized closed-loop controller. To reduce computational cost and mitigate model rollout errors, policy gradients are estimated using one-step predictions of the learned dynamics rather than multi-step propagation. This leads to an online mini-batch policy gradient framework that enables policy improvement from streamed interaction data. The proposed framework is evaluated on several simulated nonlinear control benchmarks and two real-world hardware platforms, including a Kinova Gen3 robotic arm and a Unitree Go1 quadruped. Experimental results demonstrate improved sample efficiency over model-free RL baselines, superior control performance relative to model-based RL baselines, and control performance comparable to classical model-based methods that rely on exact system dynamics.
\end{abstract}

\section{Introduction}
\label{sec:introduction}
Accurate dynamical models are critical for high-performance robotic control, including manipulation, legged locomotion, and aerial systems. Classical model-based optimal control methods, such as model predictive control (MPC) \cite{anderson2007optimal, kouvaritakis2016model}, can achieve strong performance when system dynamics are known with sufficient accuracy. However, in many robotic applications, obtaining accurate models remains challenging due to nonlinear dynamics, underactuation, contact interactions, actuator uncertainties, and environment-dependent variability \cite{hutter2016anymal}. These challenges have motivated the development of model-based reinforcement learning (MBRL), in which system dynamics are learned from data and leveraged to improve control policies \cite{doya2000reinforcement}. Compared with model-free RL, MBRL can achieve significantly higher sample efficiency, which is particularly important for real-world robotic systems. However, many existing MBRL approaches rely on nonlinear deep neural networks (DNNs) for dynamics learning \cite{hernandaz1990neural,miller1990real,draeger1995model,levine2016end,deisenroth2011learning,chua2018deep,askari2025model}. While flexible, such models introduce two key challenges: (i) the computational burden of repeated multi-step nonlinear propagation during policy optimization and planning, and (ii) the potential accumulation of model errors over long-horizon rollouts, which can degrade the accuracy of policy gradient estimates \cite{tanaka1996approach, polydoros2017survey}.

An alternative direction within MBRL is to approximate nonlinear dynamics through linear evolution in a lifted space based on Koopman operator theory \cite{koopman1931hamiltonian}. By mapping system states into a higher-dimensional latent space, the underlying nonlinear dynamics can be approximately represented by a linear system \cite{mezic2015applications}. Classical approaches such as dynamic mode decomposition and extended dynamic mode decomposition construct such lifted representations by manually selecting lifting functions \cite{proctor2016dynamic, mauroy2016linear, korda2018convergence, proctor2018generalizing}. However, the choice of suitable lifting functions remains an open challenge, and the resulting high-dimensional lifted space can limit real-time applicability.

To address these limitations, recent work has explored data-driven approaches for learning Koopman representations. For example, \cite{lusch2017data} proposed using deep learning to approximate Koopman eigenfunctions, while \cite{morton2018deep, yeung2019learning, dk2, dk, bevanda2021koopmanizingflows} introduced DNNs to parameterize lifting functions and learn them directly from data, referred to as deep Koopman operator (DKO) methods in the remainder of this paper. Extensions have also been developed to time-varying nonlinear systems \cite{zhang2019online, hao2024deep} and hybrid systems with contact dynamics \cite{o2025koopman}. Despite these advances, most Koopman-based control methods combine the learned lifted dynamics with classical model-based approaches that require multi-step trajectory propagation, such as MPC \cite{korda2018linear}. Although the linear latent representation improves computational tractability, repeated state propagation, whether over a finite horizon in MPC or MBRL, may introduce cumulative errors in cost evaluation.

Motivated by these challenges, this paper proposes an online MBRL framework that integrates linear DKO-based dynamics learning with actor–critic policy optimization. The key idea is to use the learned lifted linear dynamics for one-step policy gradient estimation, rather than relying on multi-step state rollouts. Specifically, the framework simultaneously learns a linear DKO model to approximate the unknown dynamics, estimates the cost-to-go by a critic based on one-step predictions in the lifted space, and updates the policy accordingly. This results in an online learning scheme that retains the benefits of linear dynamics while mitigating error accumulation from long-horizon model propagation. The main contributions of this paper are summarized as follows:
\begin{itemize}
\item We develop an online MBRL framework that jointly learns a \emph{linear Koopman model} of the \emph{nonlinear robotic system}, along with a critic and a policy, using replayed transition data tuples for efficient training.

\item We derive a policy gradient estimator that exploits the linearity of the lifted dynamics for efficient gradient computation, while using one-step state prediction to mitigate rollout-induced errors.
    
\item We validate the proposed method on multiple simulation environments, including the pendulum, surface vehicle, linear time-invariant system, lunar lander, and bipedal walker, as well as on two hardware platforms, a Kinova Gen3 robotic arm and a Unitree Go1 quadruped. Experimental results demonstrate improved convergence rates compared with model-free RL baselines, superior control performance relative to MBRL baselines, and control performance comparable to classical model-based control using exact system dynamics.
\end{itemize}
The remainder of the paper is organized as follows. Section~\ref{ProF} formulates the problem. Section~\ref{proposedalg} presents the proposed framework. Section~\ref{NSim} presents the simulation results. Section~\ref{robotics} reports the hardware experiments. Finally, Sections~\ref{disc} and~\ref{Conc} provide the discussion and conclusion, respectively.

\textbf{\emph{Notations.}} The Euclidean norm is denoted by $\|\cdot\|$. Given a matrix $A\in\mathbb{R}^{n\times m}$, $\parallel A \parallel_F$ is its Frobenius norm, $A'$ denotes its transpose, and $A^\dagger$ represents its Moore-Penrose pseudoinverse. For a twice differentiable function $\boldsymbol{f}(\mathbf{x},\mathbf{y})$,
$\nabla_{\mathbf{x}}\boldsymbol{f}(\mathbf{x}_k)
\coloneqq 
\frac{\partial \boldsymbol{f}(\mathbf{x},\mathbf{y})}{\partial \mathbf{x}}
\big|_{\mathbf{x}=\mathbf{x}_k}$
and
$\nabla_{\mathbf{x}\mathbf{x}}\boldsymbol{f}(\mathbf{x}_k)
\coloneqq 
\frac{\partial^2 \boldsymbol{f}(\mathbf{x},\mathbf{y})}{\partial \mathbf{x}\partial \mathbf{x}}
\big|_{\mathbf{x}=\mathbf{x}_k}$
denote the first-order and second-order partial derivatives of
$\boldsymbol{f}(\mathbf{x},\mathbf{y})$ with respect to $\mathbf{x}$ evaluated at $\mathbf{x}_k$, respectively. 
Similarly,
$\nabla_{\mathbf{x}\mathbf{y}}\boldsymbol{f}(\mathbf{x}_k,\mathbf{y}_k)
\coloneqq 
\frac{\partial^2 \boldsymbol{f}(\mathbf{x},\mathbf{y})}{\partial \mathbf{x}\partial \mathbf{y}}
\big|_{(\mathbf{x},\mathbf{y})=(\mathbf{x}_k,\mathbf{y}_k)}$
denotes the mixed second-order partial derivative evaluated at $(\mathbf{x}_k,\mathbf{y}_k)$.

\section{The Problem}\label{ProF}
Given an arbitrary initial time index $\bar{t}\in\{0,1,2,\cdots \}$, we consider a classical optimal closed-loop control problem with unknown nonlinear dynamics:
\begin{equation}\label{eq_classic_oc}
\begin{aligned}
      \mathbf{u}^*(\mathbf{x}) = \arg\min_{\mathbf{u}(\mathbf{x})}\quad   J_{\bar{t}}(\mathbf{u}(\mathbf{x})) &= \sum_{t=\bar{t}}^{\infty} \gamma^{t-\bar{t}}  c(\mathbf{x}(t), \mathbf{u}(t))  \\
 \text{subject to:}\ \mathbf{x}(t+1) &= \boldsymbol{f}(\mathbf{x}(t),\mathbf{u}(\mathbf{x}(t))),\\ \mathbf{x}(t)&\in\mathcal{X},\quad \mathbf{x}(\bar{t}) = \mathbf{x}_{\bar{t}}\ \textrm{given}, 
\end{aligned}
\end{equation}
where $t = \bar{t}, \bar{t}+1, \bar{t}+2, \cdots$ is a discrete-time index, $\mathbf{x}(t)$ represents the system state, and $\mathbf{u}(\mathbf{x}(t)): \mathcal{X}\rightarrow\mathcal{U}$ denotes the closed-loop controller mapping. The function $c: \mathcal{X}\times\mathcal{U}\rightarrow\mathbb{R}$ denotes a bounded and Lipschitz continuous stage cost, $0<\gamma< 1$ is a discount factor. The unknown mapping $\boldsymbol{f}: \mathcal{X}\times\mathcal{U}\rightarrow\mathcal{X}$ describes the time-invariant nonlinear system dynamics and is assumed to be Lipschitz continuous. Throughout this paper, the state space $\mathcal{X}$ is assumed to be countably infinite.

Since the system dynamics $\boldsymbol{f}$ is assumed to be unknown, directly solving \eqref{eq_classic_oc} to obtain the optimal controller $\mathbf{u}^*(\cdot)$ is intractable. 
Instead, we assume that for any $\mathbf{x}(t)\in\mathcal{X}$ and $\mathbf{u}(t)\in\mathcal{U}$, the unknown dynamics $\boldsymbol{f}$ in \eqref{eq_classic_oc} can be represented by a parameterized function class \begin{equation}\label{eq_estimated_dyn}
    \mathbf{x}(t+1)
=
\boldsymbol{\hat f}(\mathbf{x}(t),\mathbf{u}(t), \boldsymbol{\theta}^{f *}),
\end{equation}
where $\boldsymbol{\hat f}(\cdot, \cdot, \boldsymbol{\theta}^{f*}) : \mathcal{X}\times\mathcal{U}\times\mathbb{R}^p\rightarrow\mathcal{X}$ is a known Lipschitz continuous function class parameterized by $\boldsymbol{\theta}^{f*}\in\mathbb{R}^p$. Furthermore, we assume that the optimal closed-loop controller can be rewritten as a given parameterized function class
\[
\mathbf{u}^*(\mathbf{x}(t))
=
\boldsymbol{\mu}(\mathbf{x}(t),\boldsymbol{\theta}^{\mu *}),
\]
where $\boldsymbol{\mu}(\cdot,\boldsymbol{\theta}^{\mu *}) : \mathcal{X}\times\mathbb{R}^q \rightarrow \mathcal{U}$ is a known Lipschitz continuous function parameterized by $\boldsymbol{\theta}^{\mu *} \in \mathbb{R}^q$, hereafter referred to as the \textit{policy}. Under these assumptions, the optimal closed-loop control problem reduces to estimating the optimal parameter vectors $\boldsymbol{\theta}^{f *}$ and $\boldsymbol{\theta}^{\mu *}$.

Given a dataset of observed state–input pairs from $\boldsymbol{f}$ in \eqref{eq_classic_oc} \[\mathcal{D} = \{(\mathbf{x}_i, \mathbf{u}_i), i=0,1,2,\cdots, N_{\mathrm{max}}\},\] the objective of this paper is to jointly estimate $\boldsymbol{\theta}^{f*}$ in \eqref{eq_estimated_dyn} and $\boldsymbol{\theta}^{\mu*}$ from the dataset $\mathcal{D}$. Specifically, for any initial state $\mathbf{x}_{\bar{t}}$ from $\mathcal{D}$, the \textbf{problem of interest} is to estimate $\boldsymbol{\theta}^{f*}$ and solve
\begin{equation}\label{eq_inf_cosfunc}
    \begin{aligned}
        \boldsymbol{\theta}^{\mu *} = \arg\min_{\boldsymbol{\theta}^\mu\in\mathbb{R}^q}\ J_{\bar{t}}(\boldsymbol{\theta}^\mu) &= \sum_{t=\bar{t}}^{\infty} \gamma^{t-\bar{t}}  c(\mathbf{x}(t), \mathbf{u}(t)) \\ 
 \text{subject to:}\ \ \mathbf{x}(t+1) &= \boldsymbol{\hat f}(\mathbf{x}(t),\mathbf{u}(t),\boldsymbol{\theta}^{f *}),  \\  \mathbf{u}(t) &= \boldsymbol{\mu}(\mathbf{x}(t), \boldsymbol{\theta}^\mu), \\  \mathbf{x}(t)&\in\mathcal{X}, \quad \mathbf{x}(\bar{t}) = \mathbf{x}_{\bar{t}}, \quad \mathbf{x}_{\bar{t}} \ \text{given}.
    \end{aligned}
\end{equation}

\section{Methodology}\label{proposedalg}
In this section, we first discuss the main challenges associated with solving \eqref{eq_inf_cosfunc}. We then present a mini-batch learning framework that addresses these challenges by coupling online data collection, DKO-based dynamics learning, critic approximation, and actor updates.

\subsection{Challenges and Main Ideas}
To solve \eqref{eq_inf_cosfunc}, existing approaches in the literature typically convert the constrained optimal control problem into an unconstrained optimization problem by incorporating the system dynamics and policy into the objective function. This formulation leads to policy gradient methods, where the policy parameters are updated using gradient descent.

Specifically, let $k = 0,1,2,\ldots$ denote the iteration index, $\boldsymbol{\theta}_k^\mu \in \mathbb{R}^q$ denote the estimate of $\boldsymbol{\theta}^{\mu*}$ at iteration $k$, and let $\alpha_k^\mu > 0$ denote the step size. Starting from a given initial parameter $\boldsymbol{\theta}_0^\mu$, the policy is updated according to
\begin{equation}\label{eq_gd_mu}
\boldsymbol{\theta}_{k+1}^{\mu}
=
\boldsymbol{\theta}_k^{\mu}
-
\alpha_k^{\mu}
\nabla_{\boldsymbol{\theta}^{\mu}}J_{\bar{t}}(\boldsymbol{\theta}_k^{\mu}),
\end{equation}
where $\nabla_{\boldsymbol{\theta}^{\mu}}J_{\bar{t}}(\boldsymbol{\theta}_k^{\mu})$ is referred to as the \emph{policy gradient}. Note here that the policy gradient in \eqref{eq_gd_mu} is derived under the assumption of perfect knowledge of the system dynamics in \eqref{eq_estimated_dyn}. In practice, however, the true dynamics are unknown and must be learned concurrently with the policy. In model-based policy gradient methods, this leads to evaluating the objective in \eqref{eq_inf_cosfunc} by propagating learned dynamics over multiple steps. 

Let $\boldsymbol{\theta}_k^f \in \mathbb{R}^p$ denote the estimate of $\boldsymbol{\theta}^{f*}$ at iteration $k$. The mismatch between the learned and true dynamics induces the following stage cost estimation error:
\begin{equation}
\begin{aligned}
\boldsymbol{\epsilon}_k^{\mathrm{c}}(t+1)
&=
c(\boldsymbol{\hat f}(\mathbf{x}(t),\mathbf{u}(t),\boldsymbol{\theta}^{f*}),\, \mathbf{u}(t+1)) \\
&\quad -
c(\boldsymbol{\hat f}(\mathbf{x}(t),\mathbf{u}(t),\boldsymbol{\theta}_k^f),\, \mathbf{u}(t+1)). \nonumber
\end{aligned}
\end{equation}
As a result, evaluating the objective function in \eqref{eq_inf_cosfunc} using $\boldsymbol{\theta}_k^f$ leads to the cumulative error
\begin{equation}
J_{\bar{t}}(\boldsymbol{\theta}_k^f,\boldsymbol{\theta}_k^\mu)
=
J_{\bar{t}}(\boldsymbol{\theta}^\mu) + 
\sum_{t=\bar{t}+1}^{\infty}\boldsymbol{\epsilon}_k^c(t). \nonumber
\end{equation}
This reveals a key challenge of classic model-based policy gradient methods: the reliance on multi-step rollout of learned (typically nonlinear) dynamics can introduce cumulative errors in policy gradient estimation and increase computational cost.

To address this challenge, the proposed method introduces two key components. First, the nonlinear system dynamics are represented using a linear DKO dynamics, which enables efficient gradient computation compared to nonlinear models. Second, a critic network \cite{konda1999actor} is employed to approximate the objective function $J_{\bar{t}}(\boldsymbol{\theta}^\mu)$ in \eqref{eq_inf_cosfunc}, i.e., \begin{equation}\label{eq_critic_func}
    J_{\bar t}(\boldsymbol{\theta}^\mu)
=
V(\mathbf{x}(\bar t),\boldsymbol{\theta}^{J*}),
\end{equation}
where $V(\cdot,\boldsymbol{\theta}^{J*}):\mathbb{R}^n\times\mathbb{R}^s\rightarrow\mathbb{R}$ is assumed to be Lipschitz continuous and parameterized by $\boldsymbol{\theta}^{J*}\in\mathbb{R}^s$. This formulation enables the objective function to be computed using one-step predictions:
\begin{equation}\label{eq_key_idea_onestep}
    \begin{aligned}
        J_{\bar t}(\boldsymbol{\theta}^\mu)
&=
c(\mathbf{x}_{\bar t},
\boldsymbol{\mu}(\mathbf{x}_{\bar t},\boldsymbol{\theta}_k^\mu))
+
\gamma J_{\bar t+1}(\boldsymbol{\theta}^\mu) \\& = c(\mathbf{x}_{\bar t},
\boldsymbol{\mu}(\mathbf{x}_{\bar t},\boldsymbol{\theta}_k^\mu))
+
\gamma V(\mathbf{x}(\bar t + 1),\boldsymbol{\theta}^{J*})
    \end{aligned}
\end{equation}
thereby avoiding repeated multi-step rollout. In contrast to conventional approaches, which require long-horizon propagation to estimate gradients, the proposed framework leverages the learned critic network to replace the accumulated cost with a one-step estimate.

\begin{figure}[ht]
    \centering
\includegraphics[width=0.9\linewidth]{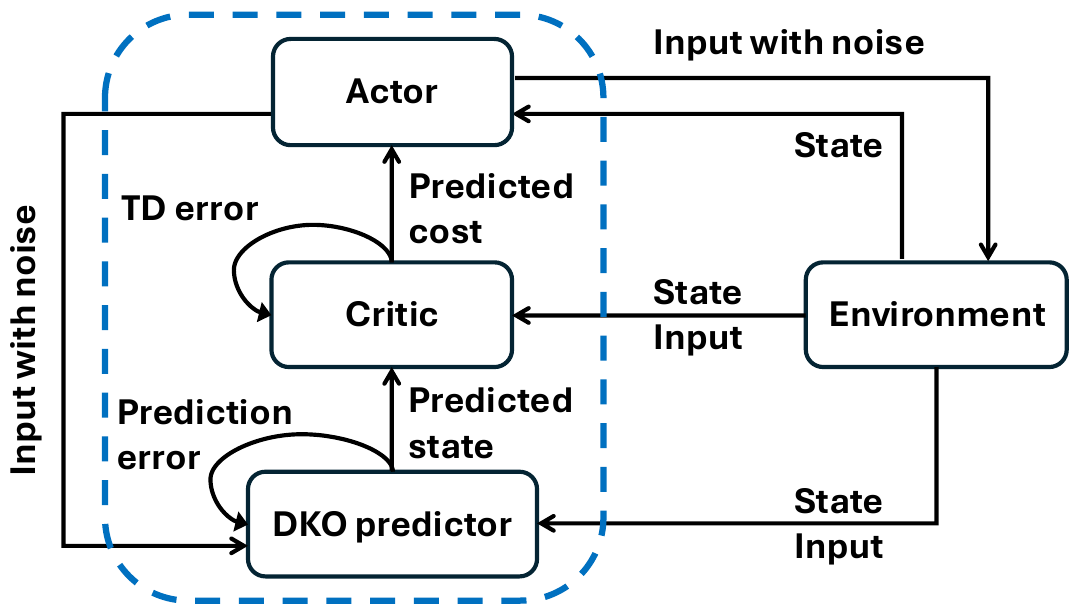}
    \caption{PGDK-Online framework.}
    \label{fig:alg_online}
\end{figure}

Together, these components reduce the impact of cumulative estimation errors in policy gradient approximation while improving computational efficiency, leading to a more stable and scalable policy optimization framework.  The overall structure is illustrated in Fig.~\ref{fig:alg_online}. 

\subsection{The Proposed Framework}
We now introduce an online framework for approximating the optimal policy parameters $\boldsymbol{\theta}^{\mu *}$ in \eqref{eq_inf_cosfunc}. The framework contains four interacting components: (i) an online data collection module, (ii) a DKO module for dynamics approximation to obtain $\boldsymbol{\theta}^{f*}$ in \eqref{eq_estimated_dyn}, (iii) a critic module for cost value approximation, and (iv) an actor module for policy optimization. The proposed framework, termed \emph{online policy gradient with deep Koopman representation} (PGDK-Online), is summarized in Algorithm~\ref{alg_pgdk}.
\begin{algorithm2e}[t]
\SetKw{Return}{Return:}
\caption{Online Policy Gradient with Deep Koopman Representation (PGDK-Online)}
\label{alg_pgdk}
\SetAlgoLined
\LinesNumbered
Initialize $\boldsymbol{\theta}_0^f \in \mathbb{R}^p$, 
$\boldsymbol{\theta}_0^J \in \mathbb{R}^s$, 
$\boldsymbol{\theta}_0^\mu \in \mathbb{R}^q$ for 
$\boldsymbol{g}(\cdot,\boldsymbol{\theta}^f)$, 
$V(\cdot,\boldsymbol{\theta}^J)$, 
and $\boldsymbol{\mu}(\cdot,\boldsymbol{\theta}^\mu)$, respectively\\
Set $k = 0$. Specify step sizes $\{\alpha_k^f\}_{k=0}^K$, $\{\alpha_k^J\}_{k=0}^K$, $\{\alpha_k^\mu\}_{k=0}^K$, discount factor $\gamma$, number of episodes $E$, horizon $T$, batch size $N$, exploration schedule $\sigma(t)$, and data memory\\
\For{$\textnormal{episode} = 1,\dots,E$}{
    Reset state $\mathbf{x}(0)=\mathbf{x}_0$; initialize noise process $W$\\
    \For{$t = 0,\dots,T$}{
        Apply control:
        $\boldsymbol{\bar u}(t) = \boldsymbol{\mu}(\mathbf{x}(t),\boldsymbol{\theta}_k^\mu) + \sigma(t) W(t)$\\
        Observe next state $\mathbf{x}_{t+1}$ and store 
        $(\mathbf{x}_t,\boldsymbol{\bar u}_t,\mathbf{x}_{t+1})$ in data memory\\
        Sample a batch of $N$ tuples from data memory for gradients computation\\
        Update $A_k$, $B_k$, $C_k$, and $\boldsymbol{\theta}_k^f$ using \eqref{eq_gd_thetaf}\\
        Update $\boldsymbol{\theta}_k^J$ and $\boldsymbol{\theta}_k^\mu$ using \eqref{TD_update} and \eqref{eq_gd_L2}, respectively\\
        $k \leftarrow k + 1$
    }
}
\Return{$\boldsymbol{\theta}_{k+1}^f$, $\boldsymbol{\theta}_{k+1}^J$, $\boldsymbol{\theta}_k^\mu$}
\end{algorithm2e}

\subsubsection{Data Collection and Data Batch Sampling}
To ensure adequate exploration of the state space $\mathcal{X}$ during online learning, the controller executes an exploration policy obtained by perturbing the current policy with a noise process \cite{lillicrap2015continuous}. Specifically, let $t_i$ denote the system time at the $i$-th observation. Given the state $\mathbf{x}(t_i)$, the executed control is
\begin{equation}\label{eq_exp_policy}
\boldsymbol{\bar u}(t_i) = \boldsymbol{\mu}(\mathbf{x}(t_i), \boldsymbol{\theta}_k^\mu) + \sigma(t_i)W(t_i)
\end{equation}
where $W(t_i)\in\mathcal{W}\subset\mathbb{R}^m$ is sampled from a noise process and $\sigma(t_i)\geq 0$ is a time-decay factor. Common choices for $W(t_i)$ include Gaussian noise or the Ornstein--Uhlenbeck process \cite{uhlenbeck1930theory}. The perturbation term promotes exploration early in training while gradually vanishing as the policy improves.

Applying the exploratory policy in \eqref{eq_exp_policy} to the unknown system dynamics \eqref{eq_classic_oc} produces the next state $\mathbf{x}(t_i+1)$. For convenience, we represent the $i$-th observed transition by the tuple
\[
(\mathbf{x}_i,\mathbf{u}_i,\mathbf{x}_i^+) := 
(\mathbf{x}(t_i), \boldsymbol{\bar u}(t_i), \mathbf{x}(t_i+1)).
\]
Since the proposed PGDK-Online operates in an online learning setting, the transition samples are collected sequentially as the system interacts with the environment. The observed tuples $(\mathbf{x}_i, \mathbf{u}_i, \mathbf{x}_i^+)$ are stored in a finite-sized data memory $\mathcal{D}_i$, which has a fixed maximum capacity $N_{\mathrm{max}}$, and is defined as follows:
\[\mathcal{D}_i = \cup_{j=1}^i\{(\mathbf{x}_j, \mathbf{u}_j, \mathbf{x}_j^+)\},\quad i\leq N_{\mathrm{max}}.\] For $i>N_{\mathrm{max}}$, the earliest $i-N_{\mathrm{max}}$ observed tuples are discarded to accommodate new data. The value of $N_{\mathrm{max}}$ is typically chosen to be large. 

To mitigate the computational burden associated with large datasets during gradient estimation, we employ mini-batch gradient descent \cite{mnih2013playing}. Specifically, at each iteration $k$, a data batch of $N$ tuples ($N\ll N_{\mathrm{max}}$) is sampled from $\mathcal{D}_i$, with its index set $\mathcal{I}_k$ drawn uniformly from all subsets of $\{1,2,\cdots,i\}$ with size $N$, provided that $i\geq N$. The sampled data batch is then used for gradient estimation, as summarized in Lines $6\sim 8$ of Algorithm~\ref{alg_pgdk}.

\subsubsection{Dynamics Approximation using Deep Koopman Operator (DKO)}
To approximate the unknown dynamics $\boldsymbol{f}$ in \eqref{eq_classic_oc}, we employ a DKO representation for \eqref{eq_estimated_dyn} given by
\begin{equation}\label{eq_approx_f}
\mathbf{x}(t+1)
=
C^*
\!\left(
A^*\boldsymbol{g}(\mathbf{x}(t),\boldsymbol{\theta}^{f*})
+
B^*\mathbf{u}(t)
\right),
\end{equation}
where $\boldsymbol{g}(\cdot,\boldsymbol{\theta}^{f*}):\mathbb{R}^n\rightarrow\mathbb{R}^r$ denotes a lifting function of Koopman operator with $r\ge n$, assumed to be Lipschitz continuous. 
The parameter $\boldsymbol{\theta}^{f*}\in\mathbb{R}^p$ and matrices
$A^*\in\mathbb{R}^{r\times r}$,
$B^*\in\mathbb{R}^{r\times m}$,
$C^*\in\mathbb{R}^{n\times r}$
are unknown constants to be learned.

The dynamics \eqref{eq_approx_f} follows from the Koopman representation
\begin{align}
\boldsymbol{g}(\mathbf{x}(t+1),\boldsymbol{\theta}^{f*})
&=
A^*\boldsymbol{g}(\mathbf{x}(t),\boldsymbol{\theta}^{f*})
+
B^*\mathbf{u}(t),
\label{eq_lift_koopman}
\\
\mathbf{x}(t+1)
&=
C^*\boldsymbol{g}(\mathbf{x}(t+1),\boldsymbol{\theta}^{f*}),
\label{eq_x_koopman}
\end{align}
where \eqref{eq_lift_koopman} describes the linear evolution of the lifted state and \eqref{eq_x_koopman} reconstructs the system state from the lifted coordinates.
We refer to \cite{hao2024deep} for a detailed discussion on the approximation properties associated with the lifting function $\boldsymbol{g}$.

One way to obtain \eqref{eq_approx_f} is to minimize the approximation error of \eqref{eq_lift_koopman}--\eqref{eq_x_koopman} over the observed transition tuples:
\begin{equation}\label{eq_L_f}
\begin{aligned}
A^*,B^*,C^*,\boldsymbol{\theta}^{f*}
=
\arg\min_{A,B,C,\boldsymbol{\theta}^f}
\sum_i
\Big(
\|
\boldsymbol{g}(\mathbf{x}_i^+,\boldsymbol{\theta}^f)
-
A\boldsymbol{g}(\mathbf{x}_i,\boldsymbol{\theta}^f)
\\-
B\mathbf{u}_i
\|^2
+
\|
\mathbf{x}_i^+
-
C\boldsymbol{g}(\mathbf{x}_i^+,\boldsymbol{\theta}^f)
\|^2
\Big),
\end{aligned}
\end{equation}
where the first norm square corresponds to the lifted-state prediction error, while the second norm square enforces reconstruction error.

Since the number of data tuples is typically large, we solve \eqref{eq_L_f} using mini-batch gradient descent \cite{mnih2013playing}. 
Let $\boldsymbol{\theta}_k^f$ denote the parameter estimate of $\boldsymbol{\theta}^{f *}$ at iteration $k$. 
Given a sampled data batch, we construct the following data matrices:
\begin{equation}\label{xyudata}
\begin{aligned}
\bX_k
&=
[\mathbf{x}_1,\ldots,\mathbf{x}_N]
\in\mathbb{R}^{n\times N},
\\
\mathbf{U}_k
&=
[\mathbf{u}_1,\ldots,\mathbf{u}_N]
\in\mathbb{R}^{m\times N},
\\
\bar{\bX}_k
&=
[\mathbf{x}_1^+,\ldots,\mathbf{x}_N^+]
\in\mathbb{R}^{n\times N},
\\
\bG_k
&=
[
\boldsymbol{g}(\mathbf{x}_1,\boldsymbol{\theta}_k^f),
\ldots,
\boldsymbol{g}(\mathbf{x}_N,\boldsymbol{\theta}_k^f)
]
\in\mathbb{R}^{r\times N},
\\
\bar{\bG}_k
&=
[
\boldsymbol{g}(\mathbf{x}_1^+,\boldsymbol{\theta}_k^f),
\ldots,
\boldsymbol{g}(\mathbf{x}_N^+,\boldsymbol{\theta}_k^f)
]
\in\mathbb{R}^{r\times N}.
\end{aligned}
\end{equation}
Using these matrices, the error loss function computed over the sampled data batch can be written compactly as
\begin{equation}
\bL_k^f
=
\frac{1}{2N}
\Big(
\|
\bar{\bG}_k
-
[A\ B]
\begin{bmatrix}
\bG_k\\
\mathbf{U}_k
\end{bmatrix}
\|_F^2
+
\|
\bar{\bX}_k
-
C\bar{\bG}_k
\|_F^2
\Big).
\end{equation}
If matrices $\bar{\bG}_k$ and
$
\begin{bmatrix}
\bG_k\\
\mathbf{U}_k
\end{bmatrix}
$
have full row rank (i.e., are right-invertible), then, starting from any given $\boldsymbol{\theta}_0^f$, the matrices and parameters are updated as follows:
\begin{equation}\label{eq_gd_thetaf}
\begin{aligned}
    [A_k\ B_k]
&=
\bar{\bG}_k
\begin{bmatrix}
\bG_k\\
\mathbf{U}_k
\end{bmatrix}^{\dagger},\\
C_k
&=
\bar{\bX}_k \bar{\bG}_k^{\dagger},\\
\boldsymbol{\theta}_{k+1}^f
&=
\boldsymbol{\theta}_k^f
-
\alpha_k^f
\nabla_{\boldsymbol{\theta}^f}
\bL_k^f(A_k,B_k,C_k,\boldsymbol{\theta}_k^f).
\end{aligned}
\end{equation}
Note that the matrices $A_k$, $B_k$, and $C_k$ are determined by the sampled data batch and $\boldsymbol{\theta}_k^f$, and are treated as constant during the update of $\boldsymbol{\theta}_k^f$. This process is summarized in Line $9$ of Algorithm~\ref{alg_pgdk}.

\subsubsection{Value Function Approximation (Critic)}
To approximate the objective function $J_{\bar t}(\boldsymbol{\theta}^\mu)$ in \eqref{eq_inf_cosfunc} using \eqref{eq_critic_func}, we estimate the parameter $\boldsymbol{\theta}^{J*}$ by minimizing the temporal-difference (TD) loss \cite{sutton1988learning} via mini-batch gradient descent, using the same data batch as in the DKO update:
\begin{equation}\label{TD_update}
\boldsymbol{\theta}_{k+1}^J
=
\boldsymbol{\theta}_k^J
-
\alpha_k^J
\nabla_{\boldsymbol{\theta}^J}
\bL_k^J(\boldsymbol{\theta}_k^J),
\quad
\boldsymbol{\theta}_0^J \text{ given},
\end{equation}
where
\begin{equation}\label{eq_td_err}
\bL_k^J(\boldsymbol{\theta}^J)
=
\frac{1}{2N}
\sum_{i\in\mathcal{I}_k}
\left(
c(\mathbf{x}_i,\mathbf{u}_i)
+
\gamma V(\mathbf{x}_i^+,\boldsymbol{\theta}^J)
-
V(\mathbf{x}_i,\boldsymbol{\theta}^J)
\right)^2. \nonumber
\end{equation}
Here, $\boldsymbol{\theta}_k^J$ denotes the estimate of $\boldsymbol{\theta}^{J*}$ at iteration $k$, and $\alpha_k^J$ is the corresponding step size. This process is summarized in Line $10$ of Algorithm~\ref{alg_pgdk}.

\subsubsection{Policy Optimization (Actor)}
To obtain the optimal policy parameter $\boldsymbol{\theta}^{\mu *}$ that generalizes across diverse initial states, the policy parameter $\boldsymbol{\theta}_k^\mu$ is updated after the dynamics and critic parameters $\boldsymbol{\theta}_k^f$ and $\boldsymbol{\theta}_k^J$ are updated according to \eqref{eq_gd_thetaf} and \eqref{TD_update}, respectively. The policy update uses the same sampled mini-batch as the DKO and critic updates, and is given by
\begin{equation}\label{eq_gd_L2}
\boldsymbol{\theta}_{k+1}^\mu
=
\boldsymbol{\theta}_k^\mu
-
\alpha_k^\mu
\nabla_{\boldsymbol{\theta}^\mu}
\hat{J}(
\boldsymbol{\theta}_{k+1}^f,
\boldsymbol{\theta}_{k+1}^J,
\boldsymbol{\theta}_k^\mu),
\quad
\boldsymbol{\theta}_0^\mu \text{ given},
\end{equation}
where
\begin{equation}\label{eq_approximate_pg}
\begin{aligned}
&\nabla_{\boldsymbol{\theta}^\mu}
\hat{J}(
\boldsymbol{\theta}_{k+1}^f,
\boldsymbol{\theta}_{k+1}^J,
\boldsymbol{\theta}_k^\mu) \\=&  \frac{1}{N}
\sum_{i \in\mathcal{I}_k} \nabla_{\boldsymbol{\theta}^\mu}\hat{J}_{i}(
\boldsymbol{\theta}_{k+1}^f,
\boldsymbol{\theta}_{k+1}^J,
\boldsymbol{\theta}_k^\mu)  \\=& \frac{1}{N} \sum_{i\in\mathcal{I}_k} \Big(\nabla_{\boldsymbol{\mu}}c(\mathbf{x}_i, \boldsymbol{\mu}(\mathbf{x}_i, \boldsymbol{\theta}_k^\mu)) \nabla_{\boldsymbol{\theta}^\mu}\boldsymbol{\mu}(\mathbf{x}_i, \boldsymbol{\theta}_k^\mu)  \\& + 
\gamma \nabla_{\boldsymbol{\hat x}}V(\boldsymbol{\hat{x}}_i^+, \boldsymbol{\theta}_{k+1}^J) \nabla_{\boldsymbol{\mu}}\boldsymbol{\hat{x}}_i^+ \nabla_{\boldsymbol{\theta}^\mu}\boldsymbol{\mu}(\mathbf{x}_i, \boldsymbol{\theta}_k^\mu)\Big)
\end{aligned}
\end{equation} with $\hat{J}_i(\boldsymbol{\theta}_{k+1}^f, \boldsymbol{\theta}_{k+1}^J, \boldsymbol{\theta}_k^\mu)$ an estimate of \eqref{eq_key_idea_onestep} at iteration $k$, and $\boldsymbol{\hat{x}}_i^+ = C_k(A_k \boldsymbol{g}(\mathbf{x}_i,\boldsymbol{\theta}_{k+1}^f) + B_k \boldsymbol{\mu}(\mathbf{x}_i, \boldsymbol{\theta}_k^\mu))$ denotes the one-step state prediction from DKO dynamics.

Note that, due to the linear DKO dynamics, $\nabla_{\boldsymbol{\mu}} \boldsymbol{\hat{x}}_i^+ = C_k B_k$ in \eqref{eq_approximate_pg} is a constant matrix, which further reduces the computational cost compared with nonlinear dynamics. The gradient in \eqref{eq_approximate_pg} serves as an approximation to the policy gradient $\nabla_{\boldsymbol{\theta}^\mu} J_{\bar t}(\boldsymbol{\theta}^\mu)$ in \eqref{eq_gd_mu}, computed using the sampled mini-batch. This procedure is summarized in Line~10 of Algorithm~\ref{alg_pgdk}.
\section{Numerical Simulations}\label{NSim}
In this section, we evaluate the proposed PGDK-Online framework on a set of benchmark control problems. The simulations are designed to assess three aspects of performance:
\begin{itemize}
    \item Convergence behavior of the learned policy compared with model-free and model-based RL baselines.
    \item Stability of the proposed PGDK-Online during learning.
    \item Control performance relative to model-free and model-based RL baselines, as well as classical model-based optimal control methods that utilize exact dynamics.
\end{itemize} To clarify the DNN architectures of PGDK-Online used for the pendulum and surface vehicle tasks, we summarize them in Table~\ref{tab:dnn_summary_pen_surface}.
\begin{table}[h]
\centering
\footnotesize
\setlength{\tabcolsep}{3pt}
\begin{tabular}{lcccc}
\toprule
Task & DNN & L1 (Act/Nodes) & L2 & Output \\
\midrule
\multirow{3}{*}{Pendulum}
& $\boldsymbol{g}$ & ReLU/400 & ReLU/300 & Linear/8 \\
& $V$ & ReLU/400 & ReLU/300 & Linear/1 \\
& $\boldsymbol{\mu}$ & ReLU/400 & ReLU/300 & Tanh/1 \\
\midrule
\multirow{3}{*}{Surface vehicle}
& $\boldsymbol{g}$ & ReLU/200 & SiLU/100 & Linear/16 \\
& $V$ & ReLU/200 & ReLU/100 & Linear/1 \\
& $\boldsymbol{\mu}$ & ReLU/200 & ReLU/100 & Tanh/2 \\
\bottomrule
\end{tabular}
\caption{DNNs architectures of the proposed PGDK-Online. Activation functions, number of nodes, and output dimensions are reported for each network.}
\label{tab:dnn_summary_pen_surface}
\end{table}

\subsection{Inverted Pendulum Balancing}
The dynamics of the inverted pendulum is given by
\begin{equation}
\ddot{\psi} = \frac{-3g}{2l}\sin(\psi + \pi) + \frac{3}{ml^2}u, \nonumber
\end{equation}
where $\psi$ denotes the angle between the pendulum and the direction of gravity, with the upright position corresponding to $\psi = 0$. The system parameters are given by $g = 10\mathrm{m/s^2}$, $m = 1\mathrm{kg}$, and $l = 1\mathrm{m}$, representing gravitational acceleration, pendulum mass, and length, respectively. The system state is defined as $\mathbf{x}(t) = [\psi(t), \dot{\psi}(t)]'$, with bounds $[-\pi, -8]' \leq \mathbf{x}(t) \leq [\pi, 8]'$. The control input $u(t)$ is a continuous scalar torque constrained within $[-2, 2]$. The system is simulated using Euler discretization:
$\psi(t+1) = \psi(t) + \dot{\psi}(t)\Delta t$ and $\dot{\psi}(t+1) = \dot{\psi}(t) + \ddot{\psi}(t)\Delta t$, where the time step is $\Delta t = 0.02\mathrm{s}$.

\subsubsection{Simulation Setup} For each simulation episode, the initial state $\mathbf{x}(0)$ is uniformly sampled between $[-\pi, -1]'$ and $[\pi, 1]'$, and each episode terminates when $t > 200$. The control objective is to stabilize the pendulum at the upright equilibrium $\mathbf{x}_{\mathrm{goal}} = [0, 0]'$. The reward function is defined as
$$r(\mathbf{x}(t), \mathbf{u}(t)) = -\Big(\psi(t)^2 + 0.1\dot{\psi}(t)^2 + 0.001u(t)^2\Big).$$
To implement the proposed PGDK-Online, the DNN architectures are summarized in Table~\ref{tab:dnn_summary_pen_surface}. All networks are trained using the Adam optimizer~\cite{kingma2014adam} with a batch size of $120$.

We compare PGDK-Online with representative model-free and model-based RL methods, as well as a classical optimal control baseline. Specifically, we consider:
\begin{itemize}
    \item Probabilistic ensembles with trajectory sampling (PETS)~\cite{chua2018deep}, a model-based RL method that learns probabilistic dynamics using DNN ensembles and performs control via MPC with a planning horizon of $25$.
    \item Soft actor--critic (SAC)~\cite{haarnoja2018soft}, a representative model-free actor--critic method.
    \item MPC using the exact system dynamics with a planning horizon of $20$.
\end{itemize}
To account for stochasticity in DNN training, all learning-based methods are evaluated over $5$ independent runs.

\subsubsection{Results and Analysis}
Table~\ref{tab:convergence_pen} summarizes the convergence behavior under both $95\%$ and $99\%$ convergence criteria. PETS achieves the fastest convergence under the stricter $99\%$ criterion, reflecting its rapid initial improvement. However, under the more practical $95\%$ criterion, PGDK-Online converges substantially faster than both PETS and SAC. This difference highlights that PETS tends to saturate early, whereas PGDK-Online maintains sustained performance improvement and achieves more reliable convergence.
\begin{table}[h]
\centering
\begin{tabular}{lcc}
\hline
Method & $95\%$ Conv. $\downarrow$ (episodes) & $99\%$ Conv. $\downarrow$ (episodes) \\
\hline
\textbf{PGDK-Online}& $\mathbf{32.4 \pm 25.6}$ & $30.0 \pm 21.5$  \\
PETS        & $78.0 \pm 0.0$  & $\mathbf{20.0 \pm 0.0}$  \\
SAC         & $67.4 \pm 11.5$ & $64.8 \pm 11.6$  \\
\hline
\end{tabular}
\caption{Convergence comparison for learning-based methods on the inverted pendulum task. Results are reported as mean $\pm$ standard deviation over $5$ independent runs.}
\label{tab:convergence_pen}
\end{table}

As reported in Table~\ref{comparison_pen}, PGDK-Online attains control performance comparable to MPC, PETS, and SAC, while requiring significantly lower computation time during deployment than online optimization-based methods such as PETS and MPC.

Overall, while MPC with exact dynamics achieves the best control performance, PGDK-Online provides the most favorable trade-off between performance and efficiency. Compared with PETS, it avoids early performance saturation and achieves more consistent convergence, and compared with SAC, it converges significantly faster. In addition, its low computational overhead makes it well-suited for real-time control applications where both performance and efficiency are critical.
\begin{table}[h]
\centering
\setlength{\tabcolsep}{3pt}
\begin{tabular}{lcccc}
\toprule
Method & Time (ms)$\downarrow$ & Avg. Step Reward$\uparrow$ & Final Err$\downarrow$  \\
\midrule
\textbf{PGDK-Online}  & $\mathbf{0.038 \pm 0.001}$ & $-0.822 \pm 0.555$ & $0.017 \pm 0.000$ \\
PETS & $154.8 \pm 4.9$ & $-0.865\pm 0.419$ & $0.021 \pm 0.012$  \\
SAC & $0.102 \pm 0.033$ & $-1.260 \pm 0.838$ & $0.011\pm 0.000$  \\
MPC (exact dyn.) & $31.2 \pm 11.5$ & $\mathbf{-0.656 \pm 0.411}$ & $\mathbf{0.000 \pm 0.000}$ \\
\bottomrule
\end{tabular}
\caption{Performance comparison of learned policies on the inverted pendulum task over $10$ initial states. Reported metrics include per-step computation time, average step reward, and final tracking error. Lower computation time, higher rewards, and lower tracking errors indicate better control performance.}
\label{comparison_pen}
\end{table}

\subsection{Goal-Tracking for Surface Vehicles}
We consider a surface vehicle modeled as a $3$-DOF rigid body with global position $\mathbf{p} = [x,\, y,\, \psi]'$ and body-frame velocity $\mathbf{v} = [u,\, v,\, r]'$, where $(x,y)$ denotes the position and $\psi$ is the yaw angle. The kinematic equations are given by
\begin{equation}
\mathbf{\dot p}
=
\begin{bmatrix}
\cos\psi & -\sin\psi & 0 \\
\sin\psi & \cos\psi & 0 \\
0 & 0 & 1
\end{bmatrix}
\mathbf{v}. \nonumber
\end{equation}
The body-frame dynamics are given by \[M\dot{\mathbf{v}} + C(\mathbf{v})\mathbf{v} + D\mathbf{v} = \boldsymbol{\tau},\]
where
\begin{equation*}
M =
\begin{bmatrix}
m & 0 & 0\\
0 & m & 0\\
0 & 0 & I_z
\end{bmatrix}, \quad
C(\mathbf{v}) =
\begin{bmatrix}
0 & 0 & -mv\\
0 & 0 & mu\\
mv & -mu & 0
\end{bmatrix},
\end{equation*}
\begin{equation*}
D =
\begin{bmatrix}
d_u & 0 & 0\\
0 & d_v & 0\\
0 & 0 & d_r
\end{bmatrix}, \quad \boldsymbol{\tau} =
\begin{bmatrix}
F_u\\
0\\
\tau_r
\end{bmatrix}
\end{equation*}
represent the inertia, Coriolis–centripetal, and damping matrices, respectively, and $\boldsymbol{\tau}$ denotes the control input vector.
 
For simulation, the system parameters are set to $m = 8.0~\mathrm{kg}$ and $I_z = 2.5~\mathrm{kg\cdot m^2}$, with linear drag coefficients $d_u = 3.0$, $d_v = 6.0$, and $d_r = 1.5$. The control inputs are the surge force $F_u \in [-12, 12]~\mathrm{N}$ and the yaw torque $\tau_r \in [-4, 4]~\mathrm{N\cdot m}$.

\subsubsection{Simulation Setup} The observation state is defined as $\mathbf{x}(t) =[
x_{\mathrm{goal}} -x(t), y_{\mathrm{goal}} -y(t), \psi_{\mathrm{goal}} -\psi(t), u(t), v(t), r(t), \cos\psi(t), \sin\psi(t), F_u(t-1), \tau_r(t-1)]'\in\mathbb{R}^8$, where $F_u$ and $\tau_r$ are initialized to zero. The control objective is to steer the boat from a given initial state to the target state $\mathbf{p}_{\mathrm{goal}} = [3.0, 4.0, 0.0]'$. For each simulation episode, the initial position $\mathbf{p}(0)$ is uniformly sampled from the range bounded by $[-0.2, -0.2, -\pi/20]'$ and $[0.2, 0.2, \pi/20]'$, and the episode terminates when $t > 200$. The reward function is defined as
\begin{equation}
\begin{aligned}
r(\mathbf{x}(t), \boldsymbol{\tau}(t))
= -0.4 \| \mathbf{p}(t) - \mathbf{p}_{\mathrm{goal}} \|^2
- 0.03 \| \mathbf{v}(t) \|^2
\\ - 0.0008 \| \boldsymbol{\tau}(t) \|^2
+ 80\, \mathbb{I}_{\mathrm{succ}}, \nonumber
\end{aligned}
\end{equation}
where $\mathbb{I}_{\mathrm{succ}}=1$ if $\parallel \mathbf{p}(t) - \mathbf{p}_{\mathrm{goal}}\parallel < 0.05$, and $\mathbb{I}_{\mathrm{succ}} = 0$ otherwise. The proposed PGDK-Online aims to learn a mapping from $(\mathbf{x}(t), \boldsymbol{\tau}(t))$ to $\mathbf{x}(t+1)$. 

For implementation of the proposed PGDK-Online, the DNN architectures are summarized in Table~\ref{tab:dnn_summary_pen_surface}. We consider SAC, PETS with a planning horizon of $25$, and MPC with a planning horizon of $20$ as baseline methods, consistent with the pendulum example. To account for randomness in DNN training, all learning-based methods are evaluated over $5$ independent runs, using a batch size of $120$.

\subsubsection{Results and Analysis}
\begin{figure}[ht]
    \centering
    \includegraphics[width=0.93\linewidth]{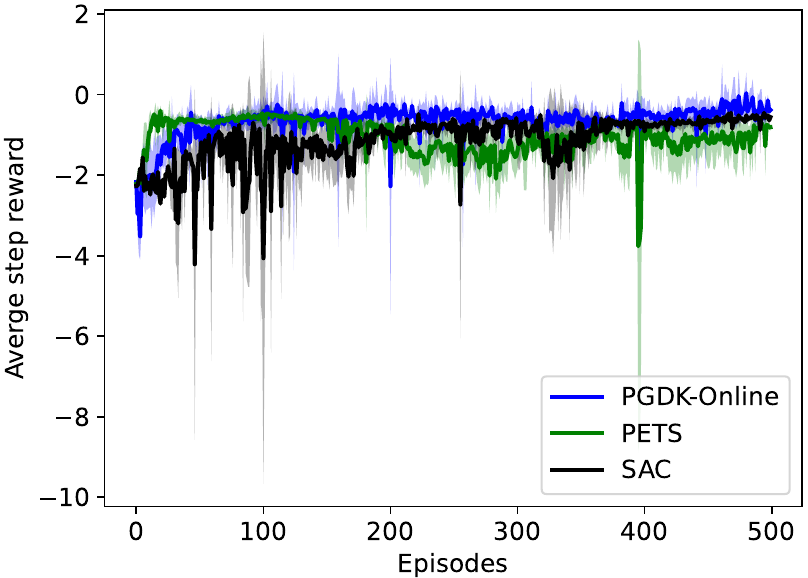}
    \caption{Learning curves of the learning-based methods on the surface vehicle task. Higher rewards indicate better control performance. Solid lines denote the mean over $5$ runs, and shaded regions represent the corresponding standard deviation.}
    \label{fig:boat_reward_his}
\end{figure}
As shown in Fig.~\ref{fig:boat_reward_his}, for the surface vehicle task, the proposed PGDK-Online achieves a convergence rate comparable to PETS, while exhibiting more stable learning behavior as the number of training episodes increases. In contrast, PGDK-Online converges significantly faster than the model-free SAC method in terms of training episodes, while achieving comparable performance. Furthermore, Table~\ref{comparison_boat} shows that the proposed PGDK-Online attains control performance comparable to MPC and outperforms PETS and SAC, while requiring substantially less computation time during deployment than the online optimization-based approaches used in PETS and MPC.
\begin{table}[t]
\centering
\setlength{\tabcolsep}{3pt}
\begin{tabular}{lcccc}
\toprule
Method & Time (ms)$\downarrow$ & Avg. Step Reward $\uparrow$ & Final Err$\downarrow$  \\
\midrule
\textbf{PGDK-Online}  & $\mathbf{0.042 \pm 0.015}$ & $0.494 \pm 0.861$ & $0.797 \pm 0.851$  \\
PETS & $172.9 \pm 0.4$ & $-0.768 \pm 0.024$ & $2.155 \pm 0.036$ \\
SAC & $0.086 \pm 0.026$ & $ -0.546 \pm 0.090$ & $1.103 \pm 0.072$  \\
MPC (exact dyn.) & $4.3 \pm 0.1$ & $\mathbf{1.025 \pm 0.102}$ & $\mathbf{0.122 \pm 0.001}$ \\
\bottomrule
\end{tabular}
\caption{Performance comparison of learned policies on the surface vehicle task over $10$ initial states. Reported metrics include per-step computation time, average step reward, and final tracking error. Lower computation time, higher rewards, and lower tracking errors indicate better control performance.}
\label{comparison_boat}
\end{table}

\subsection{Generalization Examples}
To demonstrate the generalizability of the proposed PGDK-Online, we evaluate it on both linear and nonlinear systems, including a linear time-invariant (LTI) system and two standard nonlinear benchmark environments, lunar lander and bipedal walker, from OpenAI Gym~\cite{brockman2016openai}. Furthermore, we assess control performance as system complexity increases by comparing the proposed PGDK-Online with representative model-based and model-free baselines.

\subsubsection{LTI System} The dynamics of the LTI system is \begin{equation}
    \mathbf{x}(t+1) = \begin{bmatrix}
        0.5 &0.5\\0 & 1
    \end{bmatrix} \mathbf{x}(t) + \begin{bmatrix}
        0\\1
    \end{bmatrix} u(t), \nonumber
\end{equation}
where $[-5,-5]'\leq\mathbf{x}(t)\leq [5,5]'$ and $-1\leq u(t)\leq 1$ denote the system state and control input at time $t$, respectively, and $\mathbf{x}(0)$ is uniformly sampled from the interval $[-0.1,-0.1]'$ and $[0.1,0.1]'$. The objective of this example is to design a policy for driving the system state toward the goal state $\mathbf{x}_{\mathrm{goal}} = [1,1]'$ starting from any given initial state, for which we define the following reward function: \begin{equation}
    r(\mathbf{x}(t),u(t)) = -\Big(\parallel\mathbf{x}(t)-\mathbf{x}_{\mathrm{goal}})\parallel^2 + 0.001u(t)^2 \Big). \nonumber
\end{equation} In this example, we evaluate the performance of the proposed PGDK-Online against the classical optimal control method, linear quadratic regulator (LQR), which has access to the exact system dynamics with horizon $50$ and uses the same reward function. Furthermore, we first train PGDK-Online and collect the resulting dataset, which is then used as an offline dataset to train PGDK, in order to investigate the impact of online versus offline data on policy performance.

\subsubsection{Lunar Lander} The lunar lander task requires controlling a lander to achieve a safe landing on a designated pad without any collision. The system has an $8$-dimensional state, including position, velocity, orientation, and contact indicators, and a $2$-dimensional continuous action corresponding to the main engine thrust and lateral control. We compare the proposed PGDK-Online with the model-free RL baseline deep deterministic policy gradient (DDPG)~\cite{lillicrap2015continuous}.

\subsubsection{Bipedal Walker} The bipedal walker task aims to control a bipedal robot to achieve stable forward locomotion over uneven terrain without falling. The system has a $24$-dimensional state capturing joint positions, velocities, and contact information, and $4$-dimensional continuous control inputs corresponding to joint torques. We compare the proposed PGDK-Online with DDPG.

\subsubsection{Simulation Setup} 
The DNN architectures of the proposed PGDK-Online are summarized in Table~\ref{tab:dnn_summary}. For the LTI example, we first execute the PGDK-Online algorithm to generate a replay dataset, which is then used to train the offline PGDK model. For both PGDK and LQR, the state and input are clipped to satisfy the corresponding constraints. The LQR is implemented with a horizon of $50$. For training, the batch sizes are set to $50$, $120$, and $128$ for the LTI, lunar lander, and bipedal walker tasks, respectively.
\begin{table}[h]
\centering
\footnotesize
\setlength{\tabcolsep}{3pt}
\begin{tabular}{lcccc}
\toprule
Task & DNN & L1 (Act/nodes) & L2 (Act/nodes) & Output (Act/Dim) \\
\midrule
\multirow{3}{*}{LTI}
& $\boldsymbol{g}$ & ReLU/400 & ReLU/300 & Linear/4 \\
& $V$ & ReLU/400 & ReLU/300 & Linear/1 \\
& $\boldsymbol{\mu}$ & ReLU/400 & ReLU/300 & Tanh/1 \\
\midrule
\multirow{3}{*}{Lunar}
& $\boldsymbol{g}$ & ReLU/512 & Tanh/300 & Linear/12 \\
& $V$ & ReLU/512 & ReLU/300 & Linear/1 \\
& $\boldsymbol{\mu}$ & ReLU/512 & ReLU/300 & Tanh/2 \\
\midrule
\multirow{3}{*}{Bipedal}
& $\boldsymbol{g}$ & GeLU/500 & ReLU/300 & Linear/30 \\
& $V$ & LeakyReLU/500 & LeakyReLU/300 & Linear/1 \\
& $\boldsymbol{\mu}$ & ReLU/500 & ReLU/300 & Tanh/4 \\
\bottomrule
\end{tabular}
\caption{DNN architectures of the proposed PGDK-Online. Reported are the activation functions, number of nodes, and output dimensions of the networks.}
\label{tab:dnn_summary}
\end{table}

\subsubsection{Results and Analysis.} \begin{figure*}[t]
    \centering
    \begin{subfigure}{0.33\textwidth}
        \centering
        \includegraphics[width=\linewidth]{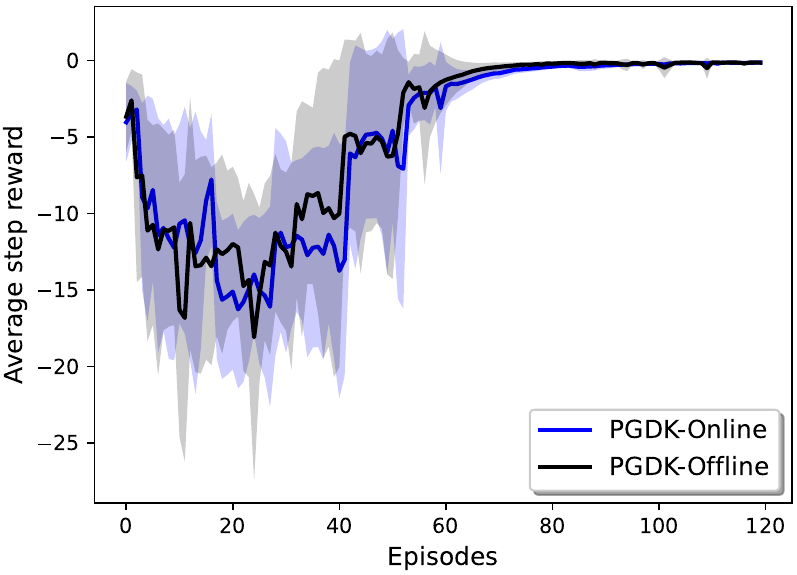}
        \caption{LTI dynamics. The solid line denotes the mean over $5$ runs, and the shaded region represents the standard deviation.}
        \label{fig:lti_train}
    \end{subfigure}
    \begin{subfigure}{0.325\textwidth}
        \centering
        \includegraphics[width=\linewidth]{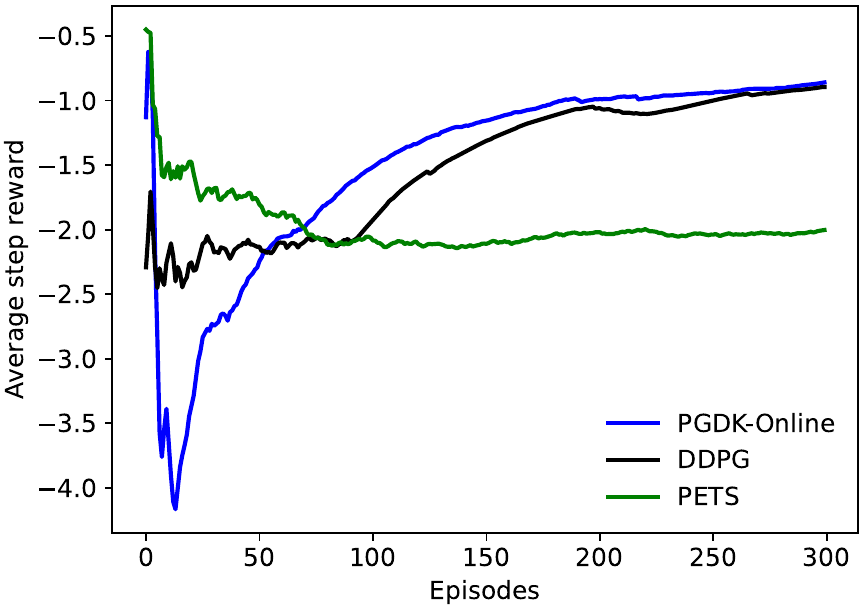}
        \caption{Cumulative average step reward for the lunar lander task.}
        \label{fig:lunar}
    \end{subfigure}
    \begin{subfigure}{0.328\textwidth}
        \centering
        \includegraphics[width=\linewidth]{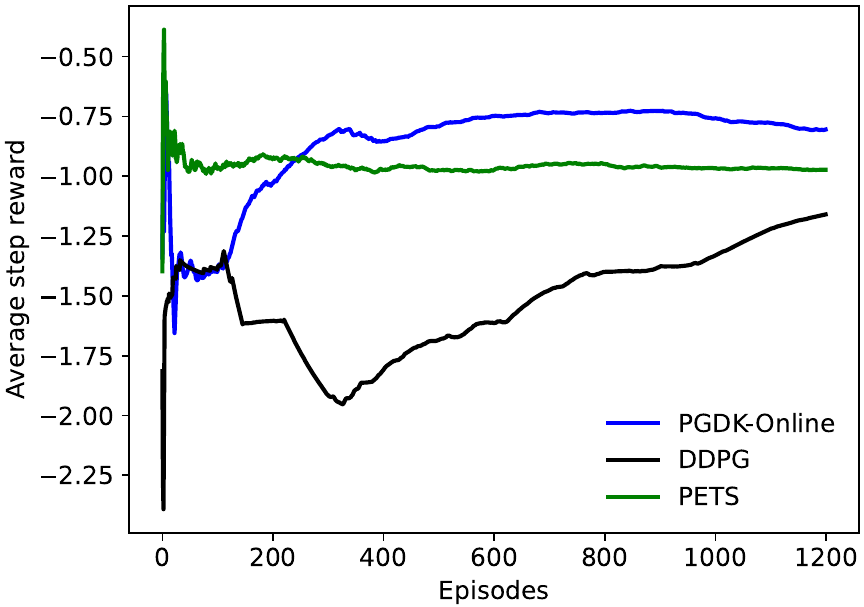}
        \caption{Cumulative average step reward for the bipedal walker task.}
    \label{fig:bipedal}
    \end{subfigure}
    \caption{Learning curves for the benchmark tasks, where a higher reward indicates better control performance.}
\end{figure*}
\begin{figure}[h]
    \centering
\includegraphics[width=0.9\linewidth]{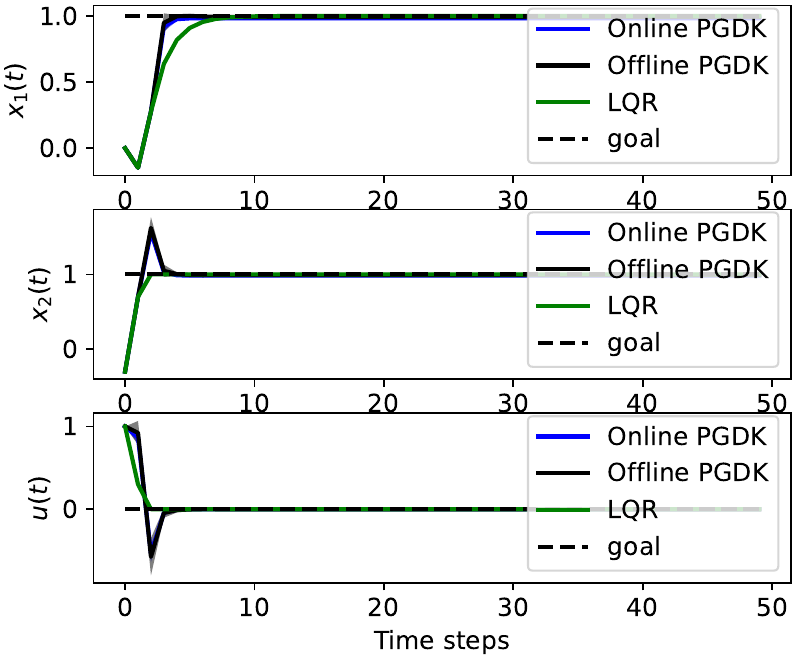}
    \caption{Trajectories from PGDK and LQR.}
    \label{fig:LTI_traj}
\end{figure}
As shown in Fig.~\ref{fig:lti_train}, the proposed PGDK-Online achieves a convergence rate comparable to that of the offline PGDK approach in terms of training episodes, while offering improved computational efficiency in gradient evaluation, since offline PGDK computes gradients using the entire dataset. Furthermore, Table~\ref{tab:general_summary} shows that the policy learned by PGDK-Online yields rewards consistently closer to those of the benchmark LQR controller than the offline PGDK method when evaluated from identical initial states. This improvement can be attributed to the reduced susceptibility of PGDK-Online to local minima. In contrast to offline PGDK, which is trained on a fixed dataset, PGDK-Online updates gradients using data batches sampled from a continually refreshed replay buffer, thereby improving exploration of the parameter space.

Notably, Fig.~\ref{fig:LTI_traj} illustrates example trajectory comparisons between the PGDK-based controllers and the LQR controller. Both online and offline PGDK tend to apply more aggressive control inputs than LQR in this LTI setting, resulting in faster convergence to the goal state at the expense of higher control effort. In contrast, the LQR controller produces smoother trajectories with more conservative inputs, reflecting its optimality under a quadratic cost formulation.

As shown in Figs.~\ref{fig:lunar}--\ref{fig:bipedal}, the proposed PGDK-Online exhibits faster convergence than PETS and DDPG on both the lunar lander and bipedal walker tasks. Table~\ref{tab:general_summary} further shows that it achieves a higher final reward than DDPG on the lunar lander task and a higher average reward on the bipedal walker task. As illustrated in Fig.~\ref{fig:lunar}, PGDK-Online experiences a sharp reward decrease during the initial $0\sim 15$ episodes, due to combined dynamics and critic learning errors in the early stage. In contrast, DDPG is affected only by critic approximation errors, while PETS is primarily limited by model inaccuracies.

\begin{table}[t]
\centering
\footnotesize
\setlength{\tabcolsep}{2.5pt}
\begin{tabular}{lcccc}
\toprule
Task & Method & Time ($\times10^{-5}$) $\downarrow$ & Avg. Reward $\uparrow$ & Final Rew. $\uparrow$ \\
\midrule
\multirow{3}{*}{LTI}
& \textbf{PGDK-Online} & $\mathbf{5.1\pm4.8}$ & $-0.12\pm0.14$ & $\mathbf{0.00}$ \\
& PGDK-Off & $5.2\pm4.5$ & $-0.16\pm0.14$ & $-0.02$ \\
& LQR (exact dyn.) & $1.6\mathrm{e}{5}$ & $\mathbf{-0.06\pm 0.07}$ & $\mathbf{0.00}$ \\
\midrule
\multirow{2}{*}{Lunar}
& \textbf{PGDK-Online} & $3.9\pm0.1$ & $0.13\pm0.03$ & $\mathbf{0.35\pm 0.00}$ \\
& DDPG & $3.9\pm1.4$ & $\mathbf{0.89\pm0.37}$ & $0.13\pm0.09$ \\
\midrule
\multirow{2}{*}{Bipedal}
& \textbf{PGDK-Online} & $4.1\pm0.5$ & $\mathbf{0.04\pm0.17}$ & -- \\
& DDPG & $\mathbf{3.7\pm0.4}$ & $-0.69\pm0.43$ & -- \\
\bottomrule
\end{tabular}
\caption{Performance on additional benchmarks (mean $\pm$ std over 10 episodes). Metrics include per-step computation time, average reward, and final reward. Lower computation time and higher rewards indicate better control performance.}
\label{tab:general_summary}
\end{table}

\section{Hardware Experiments}\label{robotics}
In this section, we evaluate the proposed framework on two hardware platforms, a robotic arm and a quadruped robot, to assess its applicability in real-world hardware settings. For clarity, the DNN architectures employed in the proposed PGDK-Online method are summarized in Table~\ref{tab:dnn_summary_arm_dog}. \begin{table}[h]
\centering
\footnotesize
\setlength{\tabcolsep}{3pt}
\begin{tabular}{lccccc}
\toprule
Task & DNN & L1 (Act./nodes) & L2 & L3  & Output \\
\midrule
\multirow{3}{*}{Robotic arm}
& $\boldsymbol{g}$ & L.ReLU/512 & L.ReLU/256 & -- & Tanh/12 \\
& $V$ & L.ReLU/512 & L.ReLU/256 &L.ReLU/128 & Linear/1 \\
& $\boldsymbol{\mu}$ & L.ReLU/512 & L.ReLU/256 & -- & Tanh/6 \\
\midrule
\multirow{3}{*}{Quadruped}
& $\boldsymbol{g}$ & ReLU/256 &ReLU/256 & -- & Linear/16 \\
& $V$ & ReLU/256 & ReLU/256 & -- & Linear/1 \\
& $\boldsymbol{\mu}$ & ReLU/256 & ReLU/256 & -- & Tanh/12 \\
\bottomrule
\end{tabular}
\caption{DNN architectures of the proposed PGDK-Online. L.ReLU denotes the LeakyReLU activation function.}
\label{tab:dnn_summary_arm_dog}
\end{table}

\subsection{Robotic Arm} 
We conduct hardware experiments on goal-tracking and collision-avoidance tasks using the Kinova Gen3 platform~\cite{kinovagen3}, a 7-DOF robotic arm equipped with joint encoders and torque sensors. The PGDK-Online framework is first trained in the Kinova Gazebo simulator and then deployed on the physical robot. As a result, measurement noise on the real platform is not explicitly modeled during training.

\begin{figure}[ht]
    \centering
\includegraphics[width=0.9\linewidth]{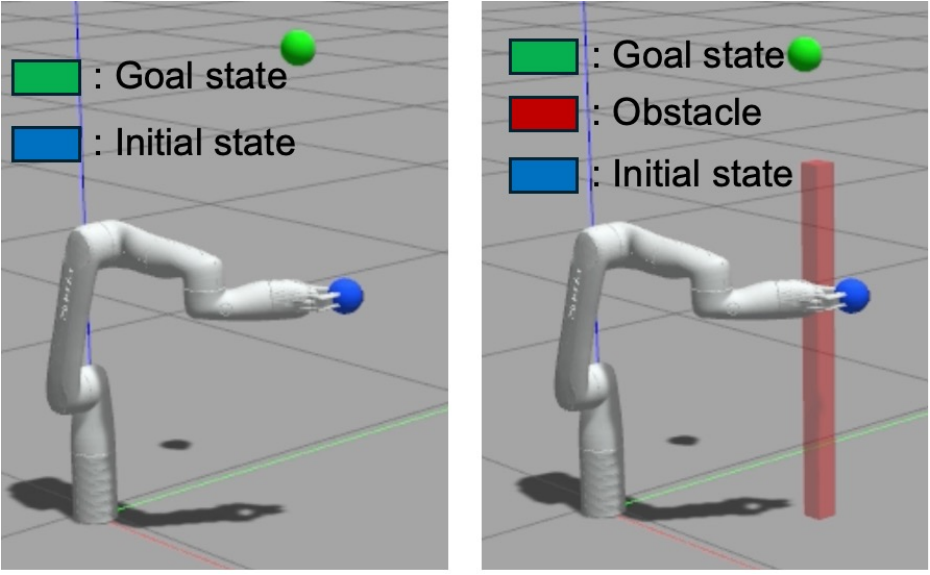}
    \caption{Gazebo simulation of the Kinova Gen3 Lite for goal-tracking tasks with and without the obstacle.}
    \label{fig:arm_env}
\end{figure}

For these tasks, which focus on goal-tracking with and without obstacle avoidance, the gripper is excluded, resulting in a 6-DOF system. The system state is defined as $\mathbf{x}(t) \in \mathbb{R}^6$, representing the joint angles of the robotic arm, initialized at $[0, \pi/4, \pi/2, 0, \pi/2]$. Due to the lack of direct torque control in the Kinova Gen3 Lite simulator, the control input is defined as the joint velocity $\mathbf{u}(t) = \dot{\mathbf{x}}(t) \in \mathbb{R}^6$. Data tuples $(\mathbf{x}_t, \mathbf{u}_t, \mathbf{x}_t^+)$ are collected at a fixed frequency of $10 \mathrm{Hz}$. The proposed algorithm is implemented in Python and executed through the robot operating system on a desktop computer equipped with an NVIDIA $2080$ Ti GPU and $128$ GB of RAM. The Gazebo simulation environment is illustrated in Fig.~\ref{fig:arm_env}.

\subsubsection{Goal-Tracking Task}
Let $\mathbf{p}(t) \in \mathbb{R}^3$ denote the position of the end-effector of the robotic arm. The objective is to drive the end-effector to $\mathbf{p}_{\mathrm{goal}}=[0.24, 0.28, 0.83]'$ while minimizing control effort. The reward function is defined as
\begin{equation}\label{eq_reward_task1}
    r(\mathbf{p}(t), \mathbf{u}(t)) 
    = -\Big( \mathbf{e}_{\mathrm{p}}(t)' Q \mathbf{e}_{\mathrm{p}}(t) + \mathbf{u}(t)' R \mathbf{u}(t) \Big),
\end{equation}
where $\mathbf{e}_{\mathrm{p}}(t) = \mathbf{p}(t) - \mathbf{p}_{\mathrm{goal}}$ denotes the position tracking error, and $Q = \mathrm{diag}(8.0, 0.02, 0.05)$ and $R = \mathrm{diag}(0.02, 0.02, 0.02, 0.02, 0.02, 0.02)$ are diagonal weighting matrices. 

Although the reward is defined based on the end-effector position $\mathbf{p}(t)$, PGDK-Online learns the system dynamics in the joint-angle space using DKO by modeling the mapping from $(\mathbf{x}(t), \mathbf{u}(t))$ to $\mathbf{x}(t+1)$. Furthermore, both Gazebo simulations and real-world experiments on the robotic arm indicate that, although the initial state $\mathbf{x}(0)$ is nominally fixed, it exhibits slight variations across training and testing episodes.

The DNN architectures for the proposed PGDK-Online are summarized in Table~\ref{tab:dnn_summary_arm_dog}. All networks are trained using the Adam optimizer with a batch size of $100$. Each episode terminates when $t > 100$. To account for stochasticity in DNN training, the experiment is conducted over $5$ independent runs.

\begin{figure}[h]
    \centering
\includegraphics[width=0.95\linewidth]{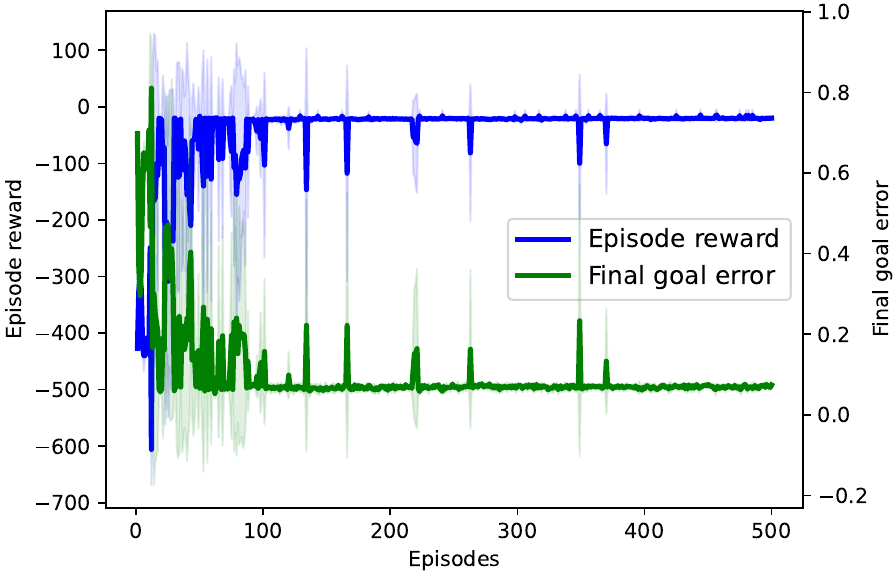}
    \caption{Learning curves and final goal-tracking errors over training episodes for the goal-tracking task. Higher rewards and lower goal-tracking errors indicate better control performance. Solid lines denote the mean over $5$ runs, and shaded regions represent the corresponding standard deviation.}
    \label{fig:arm_task1}
\end{figure}

\textbf{Results and Analysis.}
As shown in Fig.~\ref{fig:arm_task1}, the proposed PGDK-Online identifies a feasible policy within approximately $10$ episodes, with both the episode reward improving rapidly and the goal-tracking error decreasing significantly. The learning curves converge around $100$ episodes, after which performance stabilizes. The variance across runs is high during early exploration but diminishes over time, indicating improved training stability.

\begin{figure*}
    \centering
    \includegraphics[width=0.92\linewidth]{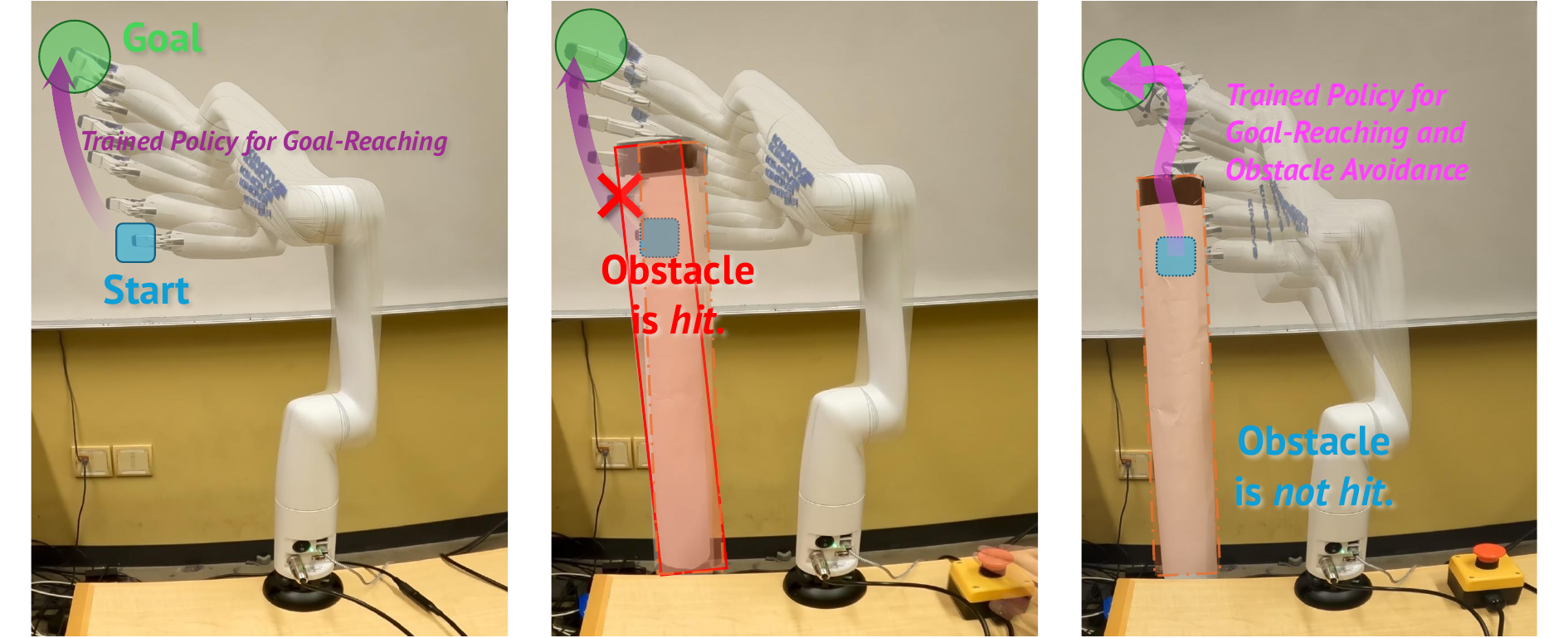}
    \caption{Experiments on the Kinova Gen3 Lite robotic arm. The task is performed with a fixed initial state (blue rounded rectangle, dashed to indicate occlusion) and a common target state (green circle). The first subfigure shows the end-effector trajectory for goal tracking generated by the policy learned by PGDK-Online. The second subfigure illustrates that, without training, the nominal trajectory collides with the obstacle (dashed orange line: nominal trajectory; red line: collision). After applying PGDK-Online, the robot successfully generates an obstacle-avoiding trajectory, as shown in the third subfigure.}
    \label{fig:arm_task1_exec}
\end{figure*}

Furthermore, Fig.~\ref{fig:arm_task1_exec} shows that the policy learned in the Gazebo simulation generalizes well to the real robotic arm, despite the presence of unknown execution noise.

\subsubsection{Goal-Tracking Task with Collision Avoidance}
To further evaluate safety-aware performance, we consider a goal-tracking task in the presence of obstacles. The learned policy from the obstacle-free goal-tracking task is used as the initial policy. In our experiments, obstacles are deliberately placed to obstruct the nominal trajectory of the learned policy.

The obstacle is modeled as a cuboid with dimensions $0.04\mathrm{m} \times 0.04\mathrm{m} \times 0.65\mathrm{m}$, located at  $[x,y,z] = [0.29,\ 0.26,\ 0]$. It forms a vertical column that blocks the direct path to the goal, requiring collision-free maneuvering. A collision penalty is incorporated into the reward function \eqref{eq_reward_task1} and is defined as
\begin{equation}\label{eq_reward_task2}
    \hat{r}(\mathbf{p}(t), \mathbf{u}(t)) 
    = r(\mathbf{p}(t), \mathbf{u}(t)) - 100 \, \mathbb{I}_{\mathrm{coll}} + 20\,\mathbb{I}_{\mathrm{succ}}, 
\end{equation}
where $\mathbb{I}_{\mathrm{coll}}$ is an indicator function that equals $1$ if a collision is detected and $0$ otherwise, with collisions determined through geometric constraints by checking intersections between the robot links and the obstacle volume. Similarly, $\mathbb{I}_{\mathrm{succ}}$ equals $1$ if $\parallel\mathbf{e}_{\mathrm{p}}(t)\parallel^2 \leq 0.05$ and $0$ otherwise.

The DNN architectures used in the proposed PGDK-Online are detailed in Table~\ref{tab:dnn_summary_arm_dog}. Training is performed using the Adam optimizer with a batch size of $100$. Each episode terminates upon collision, i.e., when $\mathbb{I}_{\mathrm{coll}} = 1$, or when the time step exceeds $t > 150$. To reduce the impact of stochasticity in DNN training, the experiment is conducted for $5$ runs.

\begin{figure}[ht]
    \centering
\includegraphics[width=0.9\linewidth]{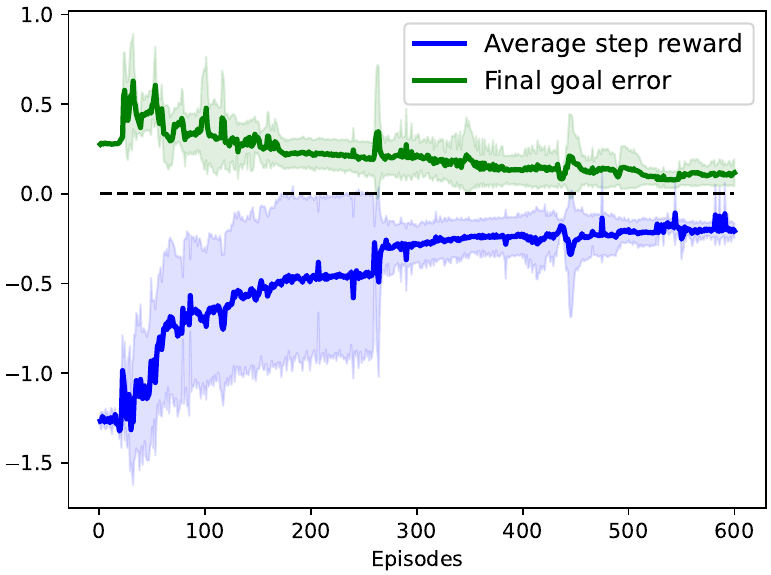}
    \caption{Learning curves and final goal-tracking errors for the goal-tracking task with obstacle avoidance. Higher rewards and lower goal-tracking errors indicate better control performance. Solid lines show the mean over $5$ runs, with shaded regions indicating standard deviation.}
    \label{fig:arm_task2}
\end{figure}

\textbf{Results and Analysis.}
As shown in Fig.~\ref{fig:arm_task2}, for the goal-tracking task with obstacle avoidance, the proposed PGDK-Online gradually improves policy performance, as indicated by increasing average step reward and decreasing goal-tracking error. Compared with the obstacle-free case, convergence is slower due to the added safety constraints, with stabilization occurring after approximately $400\sim 500$ episodes. The larger number of training episodes relative to the goal-tracking task is primarily due to early episode termination upon collision. The initially high variance reflects exploration under collision constraints, while the discrete reward function \eqref{eq_reward_task2} further increases the difficulty of critic convergence. The variance decreases over time, indicating improved training stability. Overall, the method successfully learns a collision-free policy with consistent tracking performance.

Moreover, as shown in Fig.~\ref{fig:arm_task1_exec}, starting from the policy learned in the goal-tracking task, the proposed PGDK-Online further trains the policy in the Gazebo simulator, enabling the robotic arm to reach the goal while avoiding collisions with the obstacle.

\subsection{Quadruped Robot}
We demonstrate the proposed PGDK-Online on a goal-tracking task using the Unitree Go1 quadruped robot platform~\cite{unitree_go1}, where the system states are obtained from a motion capture system. The simplified system state is defined as
\[
\mathbf{x} = [x,y,\psi,v_x,v_y,\dot{\psi}, \mathbf{h}']'\in\mathbb{R}^9,
\]
where $x$ and $y$ denote the global position, $\psi$ is the yaw angle, $v_x$ and $v_y$ are the body-frame linear velocities, and $\mathbf{h} = [l_x, l_y, a_m]' \in \mathbb{R}^3$ represents the centroidal momentum state. The body-frame velocities are related to the global-frame velocities by
$$\begin{bmatrix} v_x \\ v_y \end{bmatrix}
= \begin{bmatrix}
\cos\psi & \sin\psi\\
-\sin\psi & \cos\psi
\end{bmatrix}\begin{bmatrix} \dot{x} \\ \dot{y} \end{bmatrix}.$$
The control input is defined as
\[
\mathbf{u}=[a_{\mathrm{fwd}}, a_{\mathrm{lat}}, a_{\mathrm{z}}]'\in\mathbb{R}^3
\]
which corresponds to the commanded forward and lateral accelerations in the body frame and the desired yaw angular acceleration, respectively. The centroidal momentum dynamics are given by
\[
\mathbf{\dot h} =
\begin{bmatrix}
m\cos\psi & -m\sin\psi & 0\\
m\sin\psi & \;\;m\cos\psi & 0\\
0 & 0 & I_{zz}
\end{bmatrix}
\mathbf{u},
\]
where $m = 12~\mathrm{kg}$ is the robot mass and $I_{zz} = 0.9~\mathrm{kg\cdot m^2}$ is the yaw moment of inertia.

The control objective is to drive the quadruped to the desired goal pose $[x_{\mathrm{goal}}, y_{\mathrm{goal}}, \psi_{\mathrm{goal}}] = [1.5, 0, 0]$. As the task primarily depends on the base motion, we consider an 8-dimensional task-oriented state defined as
\begin{equation}
\label{eq_task_s}
\mathbf{s} = [\Delta x, \Delta y, \Delta \psi, \cos\psi, \sin\psi, v_x, v_y, \dot{\psi}]' \in \mathbb{R}^{8},
\end{equation}
where $\Delta x = x - x_{\mathrm{goal}}$, $\Delta y = y - y_{\mathrm{goal}}$, and $\Delta \psi = \psi - \psi_{\mathrm{goal}}$. The reward function is defined as \[r(\mathbf{s}) = -\Big(6\Delta x^2 + 20\Delta y^2 + 4.5\Delta \psi^2\Big).\] 

\subsubsection{Setup} The DNNs $\boldsymbol{g}$, $V$, and $\boldsymbol{\mu}$ follow the architectures specified in Table~\ref{tab:dnn_summary_arm_dog} and are trained using a batch size of $150$ with the Adam optimizer. Each episode terminates when $t > 200$. The PGDK-Online aims to learn a mapping from $(\mathbf{s}(t), \boldsymbol{u}(t))$ to $\mathbf{s}(t+1)$. To account for the stochasticity in DNN training, PGDK-Online is evaluated over $5$ independent runs.

\subsubsection{Results and Analysis}
\begin{figure}[ht]
    \centering
    \includegraphics[width=0.93\linewidth]{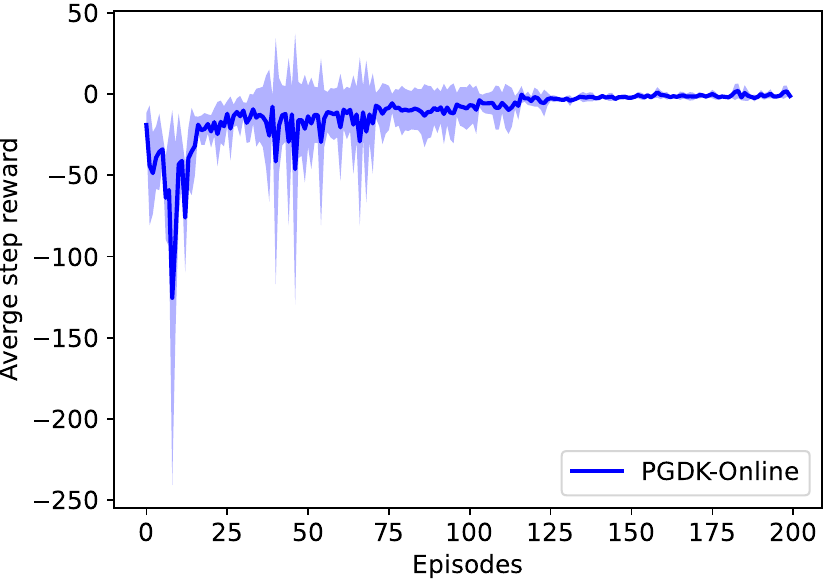}
    \caption{Learning curves. The solid line denotes the mean over $5$ runs, and the shaded region represents the standard deviation.}
    \label{fig:dog_reward_his}
\end{figure}

The learning curve in Fig.~\ref{fig:dog_reward_his} shows that PGDK-Online improves steadily over training episodes. In the early stage, the reward is low and exhibits high variance due to exploration. As training progresses, both the reward and stability improve, with the curve plateauing after approximately $100\sim 150$ episodes, indicating convergence. The reduced variance further suggests consistent performance across runs, and the final reward near zero demonstrates effective control performance. 

As shown in Table~\ref{comparison_dog}, PGDK-Online achieves significantly lower computation time than MPC, enabling real-time deployment. While MPC attains lower tracking errors, PGDK-Online achieves comparable final tracking error with only slightly higher average error. In addition, although MPC produces smoother control inputs, PGDK-Online exhibits higher control variability, reflecting a trade-off between tracking accuracy, control smoothness, and computational efficiency. Overall, PGDK-Online provides a favorable balance between control performance and real-time applicability.
\begin{table}[h]
\centering
\setlength{\tabcolsep}{3pt}
\begin{tabular}{lcccc}
\toprule
Method & Time (ms)$\downarrow$ & Avg. Err.$\downarrow$ & Final Err.$\downarrow$ & Smooth$\downarrow$  \\
\midrule
\textbf{PGDK-Online}  & $\mathbf{0.04\pm 0.01}$ & $0.72\pm 0.03$ & $0.05\pm 0.02$ & $0.98\pm 0.65$ \\
MPC (exact dyn.) & $42.3\pm 9.9$ & $\mathbf{0.57\pm 0.06}$ & $0.05\pm 0.00$& $\mathbf{0.02\pm 0.01}$ \\
\bottomrule
\end{tabular}
\caption{Performance comparison over 10 initial states in simulation. Reported metrics include per-step computation time, average and final tracking errors, and control smoothness, defined as $V = \frac{1}{T} \sum_{t=0}^{T-1} \|\mathbf{u}(t+1) - \mathbf{u}(t)\|^2$.}
\label{comparison_dog}
\end{table}

\begin{figure*}[t]
    \centering
    \includegraphics[width=0.99\linewidth]{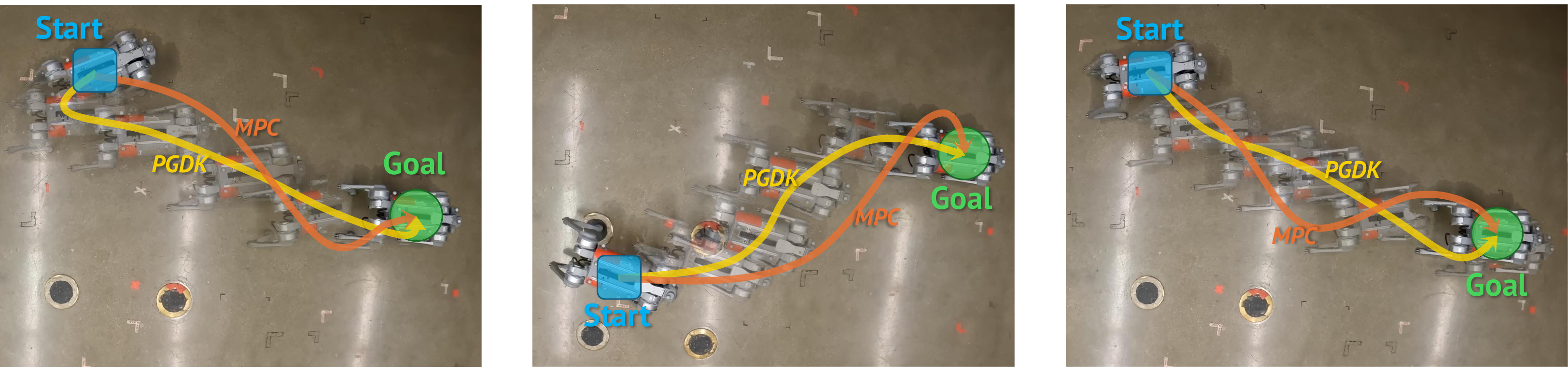}
    \caption{Experiments on the Unitree Go1 quadruped robot. Three distinct initial states (blue rounded rectangles) and a common target state (green circle) are considered. The trajectories generated by MPC with exact dynamics (orange curves with arrows) and the proposed PGDK-Online method (yellow curves with arrows) are shown.}
    \label{fig:dog_exec}
\end{figure*}

As shown in Fig.~\ref{fig:dog_exec} and Table~\ref{comparison_dog}, in simulation, MPC using exact dynamics achieves better tracking performance, as expected. In real-world execution, however, the proposed PGDK-Online attains performance comparable to MPC in terms of both average and final tracking errors. This indicates that, despite not relying on exact system dynamics, the proposed approach can achieve near-optimal performance and remains competitive under real-world uncertainties.
\begin{table}[h]
\centering
\footnotesize
\setlength{\tabcolsep}{3pt}
\begin{tabular}{lccc}
\toprule
Task & Method & Avg. Tracking Err. $\downarrow$ & Final Tracking Err. $\downarrow$ \\
\midrule
\multirow{2}{*}{Planned}
& \textbf{PGDK-Online} & $0.927 \pm 0.230$ & $\mathbf{0.032 \pm 0.009}$ \\
& MPC (exact dyn.) & $\mathbf{0.864 \pm 0.077}$ & $0.083 \pm 0.030$ \\
\midrule
\multirow{2}{*}{Executed}
& \textbf{PGDK-Online} & $\mathbf{1.063 \pm 0.160}$ & $\mathbf{0.045 \pm 0.016}$ \\
& MPC (exact dyn.) & $1.203 \pm 0.073$ & $0.112 \pm 0.039$ \\
\bottomrule
\end{tabular}
\caption{Hardware deployment performance comparison between PGDK and MPC over $3$ initial states, where both methods are evaluated from identical initial conditions. “Planned” refers to trajectories generated in the simulation environment, while “Executed” denotes real-world execution of these trajectories. Lower values indicate better tracking performance.}
\label{tab:sim2real_tracking}
\end{table}

\section{Discussion}\label{disc}
The proposed framework is motivated by the observation that learned dynamics models are most vulnerable when used for long-horizon rollouts during policy improvement. By leveraging the learned lifted dynamics primarily for one-step prediction within an actor–critic update, PGDK-Online mitigates error accumulation while still exploiting model structure to improve data efficiency.

The experimental studies are designed to evaluate this principle progressively. The pendulum example provides a controlled setting to examine convergence behavior and sample efficiency. Additional simulation benchmarks assess generalization across a broader class of nonlinear continuous-control tasks. Finally, the robotic arm and quadruped experiments demonstrate the applicability of the framework to higher-dimensional systems with practical constraints.

Several implementation aspects are critical in practice. The lifting architecture must be sufficiently expressive to ensure accurate one-step prediction. Replay buffer composition and sampled data batch size influence both convergence speed and stability. For hardware deployment, additional safety mechanisms or conservative control constraints may be required during early training. Overall, the results indicate that PGDK-Online effectively balances model utilization and robustness, making it well-suited for real-world robotic learning tasks.

\section{Concluding Remarks}\label{Conc}
In this paper, we presented PGDK-Online, an online model-based RL framework for nonlinear systems with unknown dynamics that integrates linear deep Koopman-based dynamics learning with actor–critic policy optimization. By computing policy updates using one-step lifted-space prediction, the method avoids repeated multi-step rollout of learned models, thereby improving computational efficiency and mitigating long-horizon model errors. An online mini-batch gradient descent implementation enables policy learning from streamed data, with convergence validated through numerical simulations. Extensive simulations and hardware experiments demonstrate that PGDK-Online achieves strong sample efficiency compared with model-free approaches, control performance comparable to model-based methods that rely on exact system dynamics, and significantly reduced computational cost during deployment.

Future work will focus on safety-aware online learning, improved replay buffer sampling strategies, stronger theoretical guarantees under function approximation errors, and extensions to more complex robotic scenarios, including contact-rich manipulation and multi-agent systems.



\bibliographystyle{IEEEtran}
\bibliography{refs_hao.bib}

\end{document}